\documentclass[11pt, a4paper, logo, copyright, nonumbering]{astribot}
\usepackage[authoryear, sort&compress, round]{natbib}
\usepackage{dblfloatfix}
\usepackage{ulem}
\usepackage{caption}
\usepackage{dramatist}
\usepackage{xspace}
\usepackage{pifont} 
\usepackage{multirow}
\usepackage{tcolorbox}
\usepackage{xltabular}
\usepackage{longtable}
\usepackage{hyperref}
\usepackage{amsfonts}
\usepackage{amsmath}
\usepackage{amssymb}
\usepackage{lineno}
\usepackage{adjustbox}
\usepackage{siunitx}
\usepackage{mathtools}
\usepackage[bottom]{footmisc}
\usepackage{fontawesome5}
\usepackage{subcaption}
\usepackage{setspace}
\usepackage{lipsum} 
\usepackage{multicol}
\usepackage{annotate-equations}
\usepackage{pdfpages}
\usepackage{threeparttable}
\usepackage{makecell}
\usepackage{xcolor}
\usepackage{float}

\tcbset{textmarker/.style={
        halign=center,parbox=false,boxrule=0mm,boxsep=0mm,arc=0mm,
        outer arc=0mm,left=1mm,right=1mm,top=7pt,bottom=7pt,
        toptitle=1mm,bottomtitle=1mm,oversize,}}
\newtcolorbox{hintBox}{textmarker,
    colback=yellow!10!white}
\newtcolorbox{importantBox}{textmarker,
    colback=red!10!white}
\newtcolorbox{noteBox}{textmarker,
    colback=green!10!white}

\definecolor{shadeblue}{RGB}{74, 111, 255}
\definecolor{shadered}{RGB}{255, 107, 107}
\definecolor{shadegreen}{RGB}{76, 175, 80}

\makeatletter
\def\@BTrule[#1]{%
  \ifx\longtable\undefined
    \let\@BTswitch\@BTnormal
  \else\ifx\hline\LT@hline
    \nobreak
    \let\@BTswitch\@BLTrule
  \else
     \let\@BTswitch\@BTnormal
  \fi\fi
  \global\@thisrulewidth=#1\relax
  \ifnum\@thisruleclass=\tw@\vskip\@aboverulesep\else
  \ifnum\@lastruleclass=\z@\vskip\@aboverulesep\else
  \ifnum\@lastruleclass=\@ne\vskip\doublerulesep\fi\fi\fi
  \@BTswitch}
\makeatother

\addto\extrasenglish{
}

 {\begin{list}{}%
         {\setlength{\leftmargin}{#1}}%
         \item[]%
 }
 {\end{list}}
 
\renewcommand{\today}{}

\newcommand{\model}[0]{\mbox{PHILIA}\xspace}

\title{A Glimpse into Long-term Physical Coexistence with Intelligent Robots }

\author[*]{
Astribot Team
\\
\small
\texttt{research@astribot.com}
\\
\vspace{2em}
\small
Project Page: \href{https://www.astribot.com/research/Philia}{www.astribot.com/research/Philia}
\\
\vspace{1em}
\small
Author List in \hyperref[sec:contribution]{Contributions}
}

\begin{abstract}
Long-term physical coexistence with intelligent robots requires more than capable robot policies. A persistent robotic assistant must support diverse user-facing interfaces, maintain long-horizon memory of people and preferences, coordinate across robot embodiments, and translate open-ended human intent into safe physical execution. We introduce \model, a multi-robot agent built around a robot gateway abstraction.
\model retains the rich interaction and tool ecosystem of OpenClaw, while exposing robot-local runtimes, onboard perception, navigation, speaker, and robot policies through a unified capability interface. This design decouples low-frequency, high-semantic agent reasoning from high-frequency, low-level robot execution, enabling plug-and-play integration of user interfaces, robot embodiments, and policy backends. As a result, the user experience becomes compositional: advances in user interfaces, robot embodiments, robot policies, navigation stack, or interaction algorithms can improve the overall experience without requiring a redesign of the full system. 
We validate the agentic architecture on Astribot S1 robots, while designing the robot gateway contract to support future heterogeneous robot platforms through a shared capability interface for observation, task execution, navigation, speech playback, status monitoring, and task cancellation.
We present representative use cases in which agent memory and scene understanding are grounded in physical robot actions. These use cases span interactive, open-ended household scenarios, ranging from simple organization to more challenging long-horizon and dexterous service tasks, such as packing the backpack and lifting the garbage bag. We highlight the human–robot interaction flow, where contextual understanding of user intent and preferences, together with human-in-the-loop confirmation or adjustment during execution, is essential for the robot to provide effective assistance. We hope this report serves as an invitation to imagine, explore, and shape the future of human–robot coexistence and co-creation together.
\end{abstract}

\begin{document}
\maketitle

\begin{figure}[t]
  \centering
  \includegraphics[width=\linewidth]{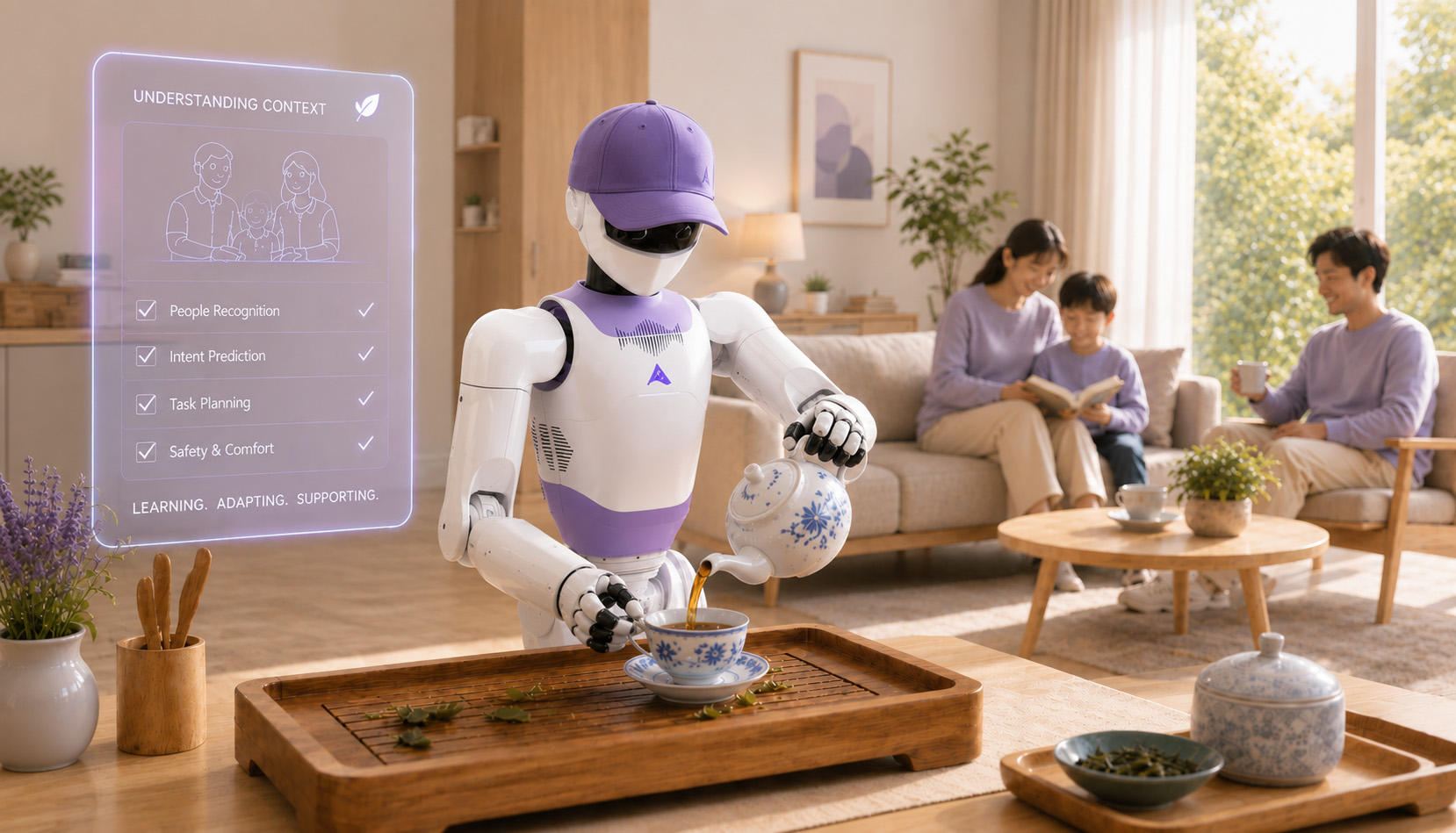}
   \label{fig:teaser}
\end{figure}

\newpage

\section{Introduction}
\label{sec:introduction}

Humans have long envisioned robots that can coexist with us in everyday environments: reliable enough to share our spaces, helpful enough to assist with open-ended requests, and socially intelligible enough to participate in daily interactions. Recent advances in both generalist and specialized robot policies have substantially expanded the capabilities of robotic systems~\citep{lumo1,brohan2022rt,zitkovich2023rt,team2024octo,kim2024openvla,black2024pi_0}. However, policy execution alone is insufficient to realize an intelligent robotic companion. Such a system must be capable of interacting through diverse interfaces, maintaining long-term memories of people and past interactions, coordinating across multiple physical embodiments, and safely translating open-ended human intentions into grounded physical actions~\citep{ahn2022saycan,driess2023palm,autort2024,wu2023tidybot}.

Most deployed robotic systems are designed around isolated command execution, where users invoke individual skills through robot-specific interfaces. While effective for standalone tasks, this paradigm does not scale to long-term human–robot coexistence or multi-robot environments that require seamless integration of heterogeneous robots, policy servers, navigation systems, user interfaces, and communication channels~\citep{hawes2017strands,liu2023smartllm,sarkar2025llamar,zhang2025llmmrs}. Moreover, robotic capabilities span fundamentally different temporal and semantic scales: language interaction, planning, and memory operate at a low frequency and high level of abstraction, whereas perception, navigation, manipulation, and safety-critical control demand high-frequency execution grounded in the physical world. We therefore advocate decoupling the robotic assistant from any specific robot embodiment, treating it as a persistent control plane responsible for user intent, memory, semantic context, tool usage, and robot orchestration, while robot-local runtimes handle perception, navigation, policy execution, low-level control, and safety-critical operations. This separation enables a unified assistant identity to coordinate multiple robot platforms while maintaining the reliability, scalability, and auditability of real-time execution~\citep{quigley2009ros,macenski2022ros2,colledanchise2018behavior}.

In this work, we introduce \model, an agentic multi-robot architecture designed for long-term physical coexistence with intelligent robots. At its core, \model adopts a robot gateway abstraction that exposes a compact set of high-level capabilities while encapsulating platform-specific middleware, policies, and control logic. Above these gateways, an OpenClaw-based control plane provides unified interaction, memory, planning, authorization, and actor-aware task routing~\citep{openclaw}. Unlike prior embodied-AI systems that primarily focus on planners, foundation models, or single-robot demonstrations~\citep{liang2023codeaspolicies,huang2023voxposer,team2024octo,kim2024openvla}, \model investigates the runtime architecture required to deploy and coordinate these components in real-world settings. Specifically, it enables robot capabilities to be exposed, invoked, monitored, interrupted, recovered, and orchestrated across multiple robot actors through a single assistant identity. By decoupling agent intelligence from robot-local execution, \model establishes a stable, auditable, and scalable control plane that makes robotic capabilities operational, composable, and transferable across heterogeneous embodiments.

This architecture enables agent intelligence to enhance robot behavior without modifying robot-local policies. Rather than encoding task semantics directly into policies, the agent provides semantic grounding, memory, and task composition, allowing the same execution backend to operate in richer contexts. For example, an agent can infer scene-specific organizing rules from observations before invoking a generic pick-and-place policy, or leverage long-term user preferences to select appropriate items when coordinating physical tasks. More broadly, the architecture supports compositional improvement: advances in user interfaces, robot embodiments, policies, perception, navigation, memory, or dialogue systems can independently enhance the overall assistant experience. Because these components interact through stable capability and gateway abstractions, improvements in any subsystem can be integrated without redesigning the entire stack, enabling scalable and continuously evolving robot assistants.

The main contributions of this work are:
\begin{itemize}
\item We propose an agent--runtime decoupled architecture that separates low-frequency semantic reasoning from high-frequency robot execution, enabling scalable, long-term operation across heterogeneous robot embodiments.

\item We introduce \textbf{robot gateways}, a capability-centric abstraction that decouples agent intelligence from robot-specific implementations, allowing semantic reasoning, memory, and planning capabilities to generalize across diverse robots without modifying robot-local execution stacks. 

\item We formulate robotic assistants as a compositional system in which advances in user interfaces, memory systems, robot embodiments, navigation modules, and policy backends can be integrated independently behind stable capability boundaries, yielding cumulative improvements in overall user experience. 

\item We validate the architecture through real-world deployments on the Astribot S1 platform and demonstrate that robot-local policies can be continuously improved while preserving the same capability contracts and system interfaces.

\end{itemize}

\section{\model}
\label{sec:model}

\subsection{Preliminaries}
\model is built around three abstractions: a persistent user-facing assistant, robot actors, and robot-local capabilities. The assistant maintains the conversational identity, long-term memory, user context, tool ecosystem, and high-level task planning. Robot actors are physical embodiments that execute the assistant's intentions through a shared capability interface, enabling heterogeneous platforms with different sensing, manipulation, navigation, or interaction capabilities to be integrated without modifying the assistant. Robot-local capabilities expose high-level operations, such as scene observation, policy execution, semantic navigation, speech output, and task interruption, while encapsulating robot-specific implementations behind a stable gateway interface.

This architecture decouples semantic reasoning from robot execution. The assistant is responsible for intent understanding, memory retrieval, tool use, and task composition, whereas robot runtimes handle real-time perception, control, policy inference, navigation, and safety-critical execution. Communication between the two is mediated through a compact gateway interface, allowing each layer to evolve independently.

Human--robot interaction follows an interleaved planning--execution paradigm. Long-horizon and compositional user requests are decomposed into plans spanning both robot and non-robot actions, with execution interleaved with perception such that intermediate observations continually inform subsequent decisions. In contrast, simple, bounded requests—including status queries, scene observations, and human interventions (e.g., stop commands or minor action adjustments)—are executed directly without explicit planning. Both execution modes share the same actor resolution, authorization, capability abstraction, and gateway interface, providing a unified interaction model across heterogeneous robot embodiments.

\begin{figure}[t]
  \centering
  \includegraphics[width=\linewidth]{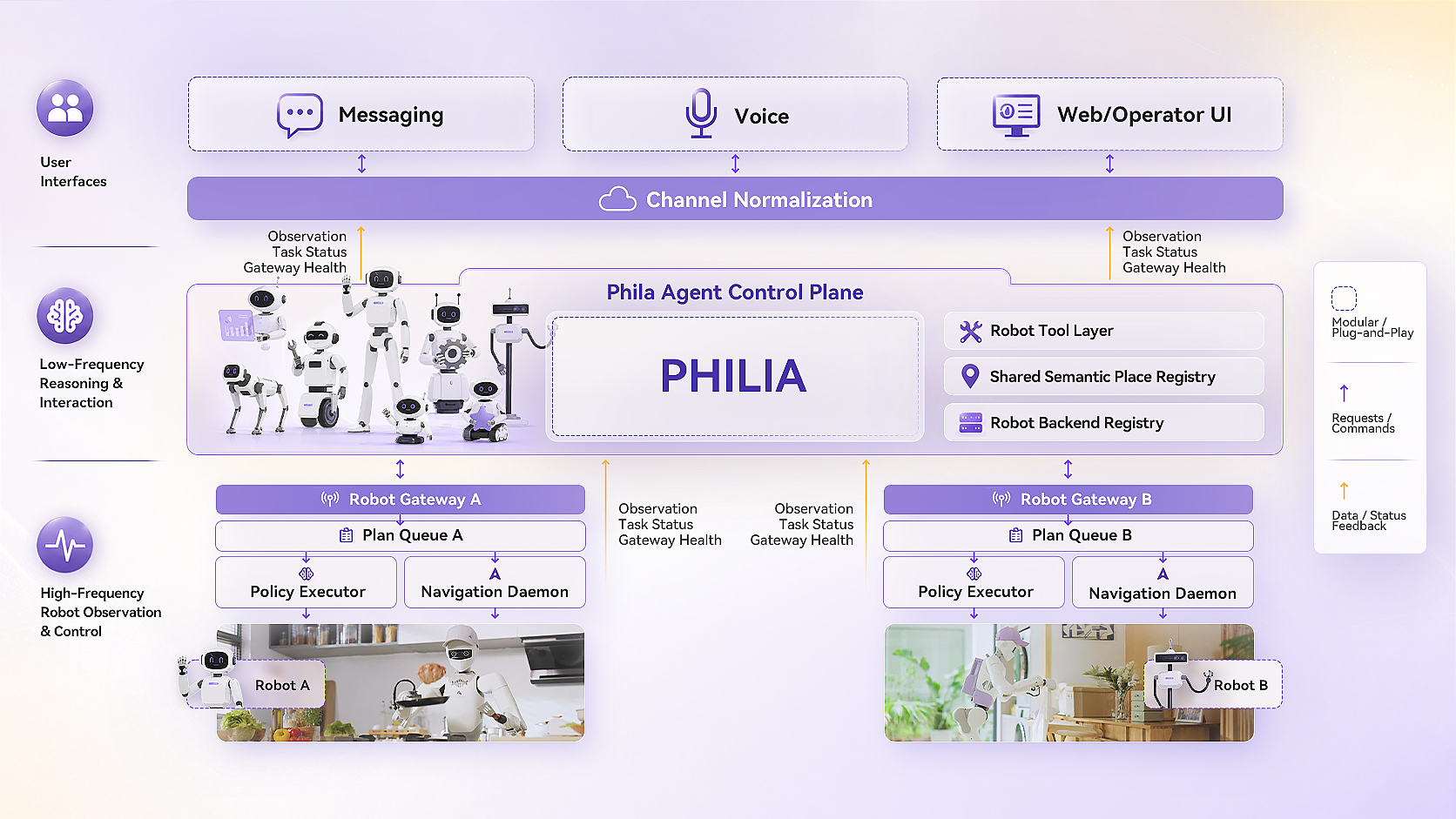}
   \caption{\textbf{System Architecture Illustration.} \model unifies user-facing interfaces, an agent control plane, and heterogeneous robot-local gateways through a shared capability abstraction.}
   \label{fig:model_architecture}
\end{figure}

\subsection{Agentic System Architecture}
\label{sec:architecture}
\paragraph{System Overview.}
As illustrated in Fig.~\ref{fig:model_architecture}, \model adopts a three-layer architecture consisting of user interfaces, an agent control plane, and robot-local gateways. User interfaces provide interaction channels, including messaging, voice, and web applications. The agent control plane maintains persistent assistant sessions, resolves target robot actors and their capabilities, enforces authorization and safety policies, and dispatches requests to the appropriate robot gateways. Robot-local gateways expose high-level capabilities while encapsulating platform-specific execution details, including middleware, sensing, navigation, bridge control, execution state, and robot policies. By cleanly separating semantic reasoning from robot-local execution, the architecture enables heterogeneous robot platforms to be integrated behind a stable capability interface, allowing new robot embodiments and policy implementations to be incorporated without modifying either the agent or the underlying robot software.

\paragraph{Compositional User Experience.}
The stable boundaries between user interfaces, the agent control plane, robot actors, and robot-local capabilities decouple system components, allowing each to evolve independently. Improvements to user interfaces, robot embodiments, reasoning and memory, perception, navigation, or robot-local policies can be incorporated without changing the surrounding system interfaces. As a result, advances at any layer directly translate into improved user experience, enabling cumulative system-wide progress through compositional integration rather than end-to-end redesign.

\paragraph{Low-frequency Reasoning and User-facing Interaction}
\model extends OpenClaw with physical embodiment while preserving its existing assistant infrastructure, including multi-modal user interfaces, tool invocation, session management, UI integration, media handling, and extensible runtime hooks. Robot interaction is treated as a first-class assistant capability rather than a separate execution mode. Within a single assistant session, the system seamlessly interleaves conversational interaction, memory retrieval, tool use, robot perception, and physical task execution through robot actors.

\paragraph{Agent Control Plane.}

The agent control plane operates entirely at the semantic level. It resolves target robot actors from user-facing names and aliases, discovers their advertised capabilities, validates capability invocations against authorization and readiness constraints, constructs execution plans for compound requests, and tracks progress across multiple actors. Free-form user instructions are interpreted through a capability-grounded interface that provides the agent with the runtime capability manifest and produces structured dispatch decisions. To balance responsiveness with robustness, request routing follows a five-stage cascade: regex fast-pass, local classifier gate, deterministic router, capability-grounded model router, and agent execution. Instead of generating low-level motor commands, the control plane emits structured capability requests that are executed by robot-local gateways. This decoupling allows semantic reasoning, planning, and long-term memory to be shared across heterogeneous robots while preserving a compact, auditable, and platform-independent execution interface.

\paragraph{Plans and Workflows.}
Bounded interactions, such as scene observation, status queries, or explicit stop commands, are dispatched directly as individual capability invocations. Compound goals are first decomposed into workflows spanning robot and non-robot actions, with each step corresponding to a capability invocation. The agent control plane schedules execution, incorporates intermediate observations to guide subsequent decisions, and handles retries, recovery, or termination based on action outcomes. Task composition therefore remains an assistant-side responsibility, while each capability is executed independently by the robot-local runtime. This separation enables flexible high-level reasoning over \textit{what} to do while delegating \textit{how} to execute each step to robot-local policies and control systems.

\paragraph{Robot Gateway Abstraction.}

A robot gateway defines the robot-local interface exposed to the assistant. Each gateway publishes a runtime manifest describing the capabilities supported by its robot, together with the corresponding input and output schemas. Because manifests are discovered dynamically, newly deployed policies or capabilities become immediately available to the agent without modifying the control plane. Behind this interface, robots may employ different middleware, SDKs, navigation systems, policy implementations, or bridge processes. The control plane remains agnostic to these implementation details, interacting only with the capabilities advertised by each gateway. The gateway interface is intentionally positioned between a low-level driver API and a full robot middleware. It does not impose a common sensor suite, action space, controller, SDK, or navigation stack. Instead, each robot advertises the capabilities it supports while encapsulating platform-specific implementation details behind the gateway. This abstraction enables a single assistant to coordinate heterogeneous robot platforms while preserving robot-specific optimization within each local runtime.

\paragraph{Unified Multi-Robot Interface.}
\model represents robot embodiments as actor-scoped instances of a shared assistant identity. An actor registry maps user-facing names to robot actors, a backend registry associates actors with robot-local gateways, and a capability catalog advertises the operations each actor supports, decoupling user interfaces, assistant logic, and robot execution. The same abstraction extends to spatial context: the control plane reasons over shared semantic place names, while each robot maintains its own gateway, navigation stack, and local map. This design enables heterogeneous robots to operate within a common semantic environment while preserving platform-specific execution, localization, and navigation. Although our current implementation targets Astribot S1 robots, additional platforms can be incorporated simply by registering new actor--gateway pairs and their capability manifests.

\paragraph{Capability Exposure.}
Manipulation policies are exposed as robot-local capabilities. The agent performs semantic reasoning, including task interpretation, policy selection, prompt generation, and coordination with perception and navigation, while the robot-local runtime handles image processing, action generation, and high-frequency control. This separation enables open-ended agent reasoning to enhance policy usability without requiring manipulation policies to incorporate interaction, memory, or task management.

\paragraph{Semantic State and Long-Horizon Context.}

State is partitioned across architectural layers according to its semantics and update frequency. The agent control plane maintains persistent semantic state, including user preferences, long-horizon plans, semantic place names, and task history, while robot-local runtimes retain high-frequency execution state, such as sensor streams, maps, coordinate frames, and policy buffers. Robot gateways exchange only lightweight semantic summaries, including capability readiness, place manifests, and task outcomes. This layered state abstraction enables the assistant to reason over persistent semantic concepts while delegating real-time perception, localization, and execution to robot-local runtimes.

\paragraph{Safety Envelope.}

Physical actions are mediated through a small set of explicit execution gates rather than relying solely on dialogue context. Before dispatch, the control plane enforces authorization, confirmation, and readiness checks based on the gateway's published state (e.g., connectivity, localization, posture, and map availability). During execution, actor-scoped task arbitration guarantees that at most one motion-changing capability is active per robot, with preemption or rejection handled according to explicit policies. Stop and cancel operations are likewise scoped to individual actors, ensuring that interruptions affect only the intended robot. These safeguards preserve the flexibility of natural-language interaction while providing a narrow, auditable, and safety-conscious path for physical execution.

\subsection{Policies as Capabilities.}
Recent advances in robot learning have significantly expanded the capability boundaries of robotic systems. We briefly summarize the learning strategies adopted to acquire and continuously improve these capabilities, training from a mixture of expert demonstrations, autonomous rollouts, and corrective interventions. During deployment, we employ training-time real-time chunking together with runtime trajectory filtering to mitigate temporal inconsistencies between training and execution. Throughout this process, action generation, action-chunk execution, and real-time control remain within the robot-local runtime, while the assistant continues to invoke the same stable capability interface as the underlying policies evolve.
\paragraph{Foundation Models as Zero-Shot Executors.} 
Recent advances in robot learning have rapidly improved generalist robot policies and large-scale cross-embodiment datasets, substantially expanding the range of tasks that robots can perform once an appropriate objective is specified. In particular, recent foundation models have begun to demonstrate promising zero-shot capabilities on simple manipulation tasks, such as picking and placing common objects, opening and closing drawers, and manipulating common household objects~\citep{lumo1}. \model is designed to seamlessly benefit from these advances by exposing robot policies through a stable capability interface, allowing improved policies to be composed with memory, user interfaces, navigation, and safety mechanisms while preserving the same assistant-facing interface for long-term autonomous assistance.

\paragraph{Task-Specific Supervised Fine-Tuning (SFT).} For tasks requiring greater dexterity or long-horizon coordination, we collect expert demonstrations to fine-tune a generalist foundation policy, yielding reliable performance within the distribution represented by the demonstrations. We refer readers to~\citep{larchenko2025task} for common supervised fine-tuning (SFT) techniques. The resulting policy provides a strong initialization for deployment, serving as both an immediately deployable controller and a stable, comparatively safe behavioral prior for subsequent real-robot data collection and iterative policy improvement. 

\paragraph{Closing the Deployment Loop.}
For high-complexity tasks, we further improve deployed policies through online or offline reinforcement learning (RL). In the offline setting, we adopt advantage-conditioned post-training to compensate for the limited coverage of failure and recovery states in demonstration-only supervised fine-tuning (SFT). Expert demonstrations rarely contain corrective behaviors for grasp failures, abnormal object configurations, motion stagnation, or repetitive local behaviors, although such states frequently arise during closed-loop execution and can lead to compounding out-of-distribution errors once the policy deviates from the demonstration manifold. Following prior intervention-based real-robot post-training paradigms~\citep{kelly2019hgdaggerinteractiveimitationlearning, intelligence2025pi06vlalearnsexperience, yu2026chi0resourceawarerobustmanipulation}, we collect autonomous rollouts from the deployed SFT policy and augment them with corrective human interventions when necessary, thereby enriching the training distribution around failure states. Rather than treating these trajectories as additional SFT data, we perform advantage-conditioned post-training by training an advantage estimator on a frame-level progress signal derived from each trajectory's normalized temporal index. The estimator predicts the relative task progress between the current and future states, and its continuous outputs are converted into task-specific binary optimality labels through percentile thresholding. The policy is then conditioned jointly on the task instruction and the optimality label, enabling it to distinguish high-utility behaviors from suboptimal alternatives under the same task semantics. This procedure performs offline policy extraction rather than online policy-gradient optimization: the SFT policy provides the underlying action manifold, rollout and intervention data expand recovery-state coverage, and advantage conditioning biases the learned policy toward behaviors with higher task utility while preserving the stability of supervised learning. 

One practical challenge in this SFT-plus-offline post-training pipeline is the mismatch between data semantics and training granularity. In our implementation, naive frame-level advantage labels are often inconsistent with 32-step action chunks: adjacent frames supervise highly overlapping future action sequences but may receive alternating positive and negative labels, introducing contradictory supervision for nearly identical action targets. To reduce this source of variance, we perform chunk-aware binarization by aggregating frame-level advantage scores within each action chunk before thresholding and, when necessary, applying temporal smoothing to suppress label oscillation. Human interventions provide valuable coverage of failure and recovery states, but are not assumed to be high-advantage samples by default. Before rollout data is incorporated into training, we analyze its statistics and visualize representative trajectories to verify that it captures reproducible recovery behaviors rather than merely increasing the diversity of noisy or suboptimal data.

\paragraph{Smooth Execution.} Contemporary robot policies generate actions in fixed-length chunks, while asynchronous inference introduces latency such that a portion of the previous chunk may already have been executed before the next prediction becomes available. If this committed action prefix is not modeled during training, consecutive chunks can become temporally misaligned, particularly in states where similar observations admit multiple plausible action continuations. Following the training-time real-time chunking formulation of~\citep{black2025trainingtimeactionconditioningefficient}, which extends earlier work on asynchronous execution and action-chunk handling~\citep{black2025realtimeexecutionactionchunking}, we sample a random inference delay during training, treat the corresponding executed prefix as conditioning, and optimize the loss only over the remaining postfix actions. The action prefix serves as a short-horizon execution context that is maintained and injected by the robot-local runtime rather than the assistant. During deployment, we further apply real-time trajectory filtering in the spirit of~\citep{gao2025humanlevelintelligencehumanlikewholebody} to improve trajectory continuity across chunk boundaries. Unlike the training objective, this runtime procedure filters and smooths consecutive action chunks without modifying the learned policy. Together, training-time real-time chunking and runtime trajectory filtering provide a temporal alignment mechanism between post-training and real-world execution: the former conditions policy prediction on committed actions, while the latter mitigates discontinuities introduced by asynchronous chunk switching. 

\subsection{Memory}
\subsubsection{Agentic Memory}
In the current system, long-term semantic memory is inherited from OpenClaw's Markdown-based memory framework. Durable user facts, preferences, and persistent instructions are stored in curated long-term memory files, while day-to-day interaction context is recorded as dated Markdown notes. These files serve as the assistant-level source of truth for persistent semantic knowledge. The use of Markdown is deliberate: it keeps memory human-readable, editable, portable, and fully inspectable, rather than embedding persistent state within opaque model parameters. To maintain efficiency, the assistant does not load the entire memory corpus into each interaction. Instead, OpenClaw provides retrieval tools that first identify relevant memory entries through semantic search and then selectively read only the required snippets. This retrieval-on-demand design preserves a compact working context while enabling previously acquired preferences, decisions, semantic locations, task history, and other long-term knowledge to be incorporated into reasoning whenever they become relevant.

\model builds upon this mechanism as a high-level bridge between long-term memory and physical action. Retrieved memories are not interpreted as direct robot commands or as the sole source of physical truth; instead, they provide semantic context for the agent control plane. For example, if a user has previously recorded a preference for sugar-free beverages or a low-calorie diet, a later request such as preparing afternoon tea can trigger retrieval of the relevant memory. The agent incorporates this information into task planning, queries the robot for the current scene, and formulates appropriate gateway-mediated perception and manipulation requests. This establishes a controlled path from memory to physical action: long-term semantic memory informs the interpretation of user intent, while the agent integrates the retrieved context with current observations, actor availability, capability manifests, and safety constraints before dispatching a structured request through the robot gateway. Robot-local runtimes remain responsible for perception, navigation, policy execution, and real-time safety checks. Consequently, memory enhances personalization and long-horizon task continuity without bypassing the gateway abstraction or allowing stored text to directly influence low-level robot control.

\subsubsection{Policy Memory}

Most existing policies model manipulation tasks as standard Markov Decision Processes (MDPs), an assumption that carries inherent limitations for long-horizon tasks reliant on temporal context. During training, the absence of sufficient sequential history as conditioning signals deprives the optimization process of structured temporal constraints, which readily leads to action space collapse and unstable policy convergence. At deployment, reliance on single-frame observations undermines the model’s ability to accurately infer the current task phase, frequently resulting in suboptimal or erroneous decision-making.

To address this limitation, \model incorporates two complementary history injection mechanisms. The first approach employs fixed-window historical action tokens, where a unified action encoder compresses and encodes the proprioceptive action sequences from the preceding 128 timesteps into compact latent representations. The second approach projects historical action trajectories onto the robot’s primary head view, embedding temporal history directly into visual observation encodings. This minimally invasive design achieves bidirectional alignment and unification of historical information across both vision tokens and language tokens.

We elaborate on the visual trajectory injection scheme as follows. We take state sequences spanning 128 timesteps prior to the current timestep \(t\) as historical input. Using the intrinsic and extrinsic parameters of the primary-view camera, we compute the historical trajectories of the left and right gripper end-effectors in the camera frame and overlay these trajectories onto the visual observation. A gradual color gradient along the rendered trajectories encodes temporal order, constructing strong visual history cues within the observation tokens.

\subsection{Navigation}
Navigation is exposed as a standard robot-local capability through the unified gateway interface, consistent with manipulation and observation. The agent reasons only over semantic places and high-level task goals, while mapping, localization, motion planning and safety validation remain fully encapsulated within the robot-local runtime. In a typical composed workflow, the robot first navigates to a target semantic place, performs local alignment to reach the policy execution pose, and then executes manipulation policies. This design keeps spatial grounding close to the robot while allowing the agent to compose navigation with interaction and memory.

\paragraph{Site Mapping and Semantic Place Annotation.} A deployment site is mapped via an operator-guided session: the robot traverses the environment while its on-board SLAM stack builds a persistent map. The operator marks a small set of semantic places (e.g., home dock, workstations) and a dedicated calibration anchor, all recorded in the persistent map frame to preserve geometric consistency across runtime restarts. The resulting place manifest is published to the control plane as named semantic references, without exposing raw map data or low-level pose state to the agent.

\paragraph{Per-Session Re-localization Calibration.} As the odometry frame is re-initialized on every navigation runtime startup and misaligned with the saved map frame, a lightweight calibration step bridges this gap. The operator drives the robot to the physical calibration anchor and triggers re-localization; the runtime captures the current odometry pose, retrieves the saved anchor pose, and computes an in-memory rigid transform between the two frames. All subsequent navigation requests apply this transform to saved place poses before planning. The agent only consumes a calibration status flag, while the transform itself remains a robot-local artifact.

\paragraph{Multi-Robot Composability.} Since semantic places are anchored in a shared map frame and re-localization runs independently per robot actor, a single mapped environment can serve multiple actors. The persistent map and place manifest are scoped to the deployment site and shared across all actors, while each robot maintains its own navigation runtime and calibration transform. Adding a new robot to an already mapped site requires only re-localization at the anchor, with no re-mapping or per-robot place re-authoring needed. This separation of map-level semantics (control plane) and live localization (robot runtime) forms the foundation of composable multi-robot navigation.

\paragraph{Local Control and Safety Boundary.} Bounded local motions, including small chassis adjustments and recovery maneuvers, are also exposed as actor-scoped gateway capabilities under the same safety envelope as other physical actions. The system does not grant the language model direct access to continuous motion control, maintaining responsiveness in daily interaction while upholding safety guardrails.

\section{Experiments}
\label{sec:experiments}

Our evaluation focuses on the control-plane dispatch boundary that bridges free-form user language and robot capabilities. Specifically, we examine whether requests are routed reliably, whether the dispatch mechanism generalizes across different language models, and how inference cost can be controlled. We regard this dispatch boundary as the central empirical component of the control-plane architecture (\S\ref{sec:model}), since every robot action—regardless of the underlying backend or manipulation policy—is ultimately mediated through it.

\subsection{Capability-Grounded Dispatch}

\paragraph{Setup.}
We evaluate routing performance on a balanced test set of $N{=}483$ utterances covering four intent classes: \texttt{robot\_task} (physical actions), \texttt{robot\_query} (current state observations), \texttt{robot\_control} (preemptive control such as stop, pause, resume, or safety override), and \texttt{non\_robot} (general dialogue and configuration). The dataset mixes formal and colloquial phrasing, Chinese and English, and explicit and implicit robot references. We compare two router configurations across seven language models of varying parameter scales: (1) a lightweight binary gate that classifies whether an utterance is robot-related; (2) a full capability-grounded router that outputs the matched capability ID, target actor, and missing context.

\paragraph{Core findings.}
As summarized in Tab.~\ref{tab:dispatch_quality}, the evaluation yields three key conclusions:
\begin{enumerate}
    \item \textbf{Model-agnostic routing reliability.} The framework delivers consistent performance across all seven tested models, from small on-device models to frontier large systems. The lightweight binary gate achieves F1 $\geq 0.96$ and recall $\geq 0.98$ on every model. The local Qwen3.5-4B completes lightweight inference in only 414 ms, making it suitable for fast pre-filtering.
    \item \textbf{High-precision full dispatch on frontier models.} For the capability-grounded configuration, frontier models reach 89--94\% top-1 skill matching accuracy on robot requests, and 83--89\% accuracy on explicit actor resolution. Claude-series models achieve a 100\% rejection rate on non-robot utterances, eliminating false-positive physical actions.
    \item \textbf{Zero-overhead skill extensibility.} The grounded router draws candidates from a runtime manifest catalog, so new skills become routable immediately after manifest publication, with no changes to the routing prompt or model required.
\end{enumerate}

\begin{table}[htbp]
\centering

\small
\setlength{\tabcolsep}{4pt}
\begin{adjustbox}{max width=\columnwidth}

\begin{tabular}{llccccr}
\hline
\textbf{Model} & \textbf{Config} & \textbf{Binary F1} & \textbf{Cap top-1} & \textbf{Non-R reject} & \textbf{Actor} & \textbf{p50 (ms)} \\
\hline
\multirow{2}{*}{gpt-5.4~\citep{openai2026gpt54}}
  & lightweight & 0.994 & --- & --- & --- & 3\,737 \\
  & grounded    & 0.984 & 0.893 & 0.991 & 0.862 & 9\,151 \\
\hline
\multirow{2}{*}{gpt-5.5~\citep{openai2026gpt55}}
  & lightweight & 0.989 & --- & --- & --- & 2\,933 \\
  & grounded    & 0.991 & 0.919 & 0.991 & 0.892 & 4\,710 \\
\hline
\multirow{2}{*}{gemini-3.5-flash~\citep{google2026gemini35flash}}
  & lightweight & 0.989 & --- & --- & --- & 6\,930 \\
  & grounded    & 0.991 & \textbf{0.939} & 0.983 & 0.877 & 10\,690 \\
\hline
\multirow{2}{*}{claude-opus-4.6~\citep{anthropic2026claudemodels}}
  & lightweight & 0.988 & --- & --- & --- & 2\,673 \\
  & grounded    & 0.981 & 0.919 & \textbf{1.000} & 0.831 & 3\,609 \\
\hline
\multirow{2}{*}{claude-opus-4.8~\citep{anthropic2026claudemodels}}
  & lightweight & 0.988 & --- & --- & --- & 2\,875 \\
  & grounded    & 0.979 & 0.904 & \textbf{1.000} & 0.831 & 3\,789 \\
\hline
\multirow{2}{*}{gpt-5.4-mini~\citep{openai2026gpt54mini}}
  & lightweight & 0.985 & --- & --- & --- & 2\,532 \\
  & grounded    & 0.989 & 0.803 & 0.974 & 0.508 & 4\,459 \\
\hline
\multirow{2}{*}{Qwen3.5-4B~\citep{qwen2026qwen35}}
  & lightweight & 0.965 & --- & --- & --- & \textbf{414} \\
  & grounded    & 0.962 & 0.883 & 0.809 & 0.277 & 10\,585 \\
\hline
\end{tabular}
\end{adjustbox}

\caption{\textbf{Dispatch quality across 7 models and 2 routing configurations.}
Lightweight configuration: binary robot/non-robot F1 for gate reliability.
Capability-grounded configuration: top-1 skill matching (198 robot cases; 115 non-robot rejection cases) and actor resolution accuracy (65 explicit mention cases).
Latency is per-call p50.}
\label{tab:dispatch_quality}
\end{table}

\subsection{Latency Decomposition and the Cascaded Routing}

Dispatch quality results motivate the cascaded design: the full capability-grounded router delivers higher precision but incurs 1.4$\times$ latency on the deployed claude-opus-4.6, and up to 25$\times$ on the smallest model, as shown in Tab.~\ref{tab:dispatch_quality}. Running the grounded router directly on a small local model yields only modest full-dispatch performance at 10.6 s latency, as it must process the complete manifest catalog and output structured results.

To balance accuracy and latency, we adopt a five-stage cascaded architecture tailored to the heavy-tailed distribution of intent complexity, with latency breakdowns in Tab.~\ref{tab:latency}. Ordered by increasing complexity: (1) regex / deterministic fast-pass, (2) local fast classifier, (3) deterministic semantic router, (4) capability-grounded semantic router, (5) end-to-end agent execution.

\begin{table}[h]
\centering
\begin{tabular}{lcccc}
\hline
\textbf{Tier} & \textbf{p50 (ms)} & \textbf{p90 (ms)} & \textbf{p99 (ms)} & \textbf{N} \\
\hline
1. Regex / deterministic fast-pass        & 1     & 1     & 1      & 5 \\
2. Local fast classifier (F1 0.965)       & 414   & 419   & 422    & 483 \\
3. Deterministic semantic router          & 1     & 1     & 1      & 5 \\
4. Capability-grounded semantic router    & 3\,609 & 4\,433 & 6\,197 & 483 \\
5. Agent turn, non-robot                  & 4\,857 & 17\,484 & 52\,327 & 2014 \\
5. Agent turn, robot tool                 & 25\,348 & 57\,264 & 288\,890 & 122 \\
\hline
\end{tabular}
\caption{\textbf{Five-tier dispatch latency.} Tiers handle intents of increasing complexity; most traffic resolves before the heavyweight grounded router.}
\label{tab:latency}
\end{table}

This hierarchical design reduces average inference cost while preserving high dispatch accuracy: most simple intents are resolved at low-latency early tiers, and only complex ambiguous requests reach the heavyweight router. Our deployed configuration uses Qwen3.5-4B for pre-filtering and claude-opus-4.6 for full routing, achieving an optimal precision-latency tradeoff: 100\% non-robot rejection at 3.6 s p50 routing latency.

\begin{figure}[H]
    \centering

    \begin{subfigure}{\linewidth}
        \centering
        \caption{Observation-grounded reasoning.}
        \includegraphics[width=\linewidth,height=0.5\textheight,keepaspectratio]{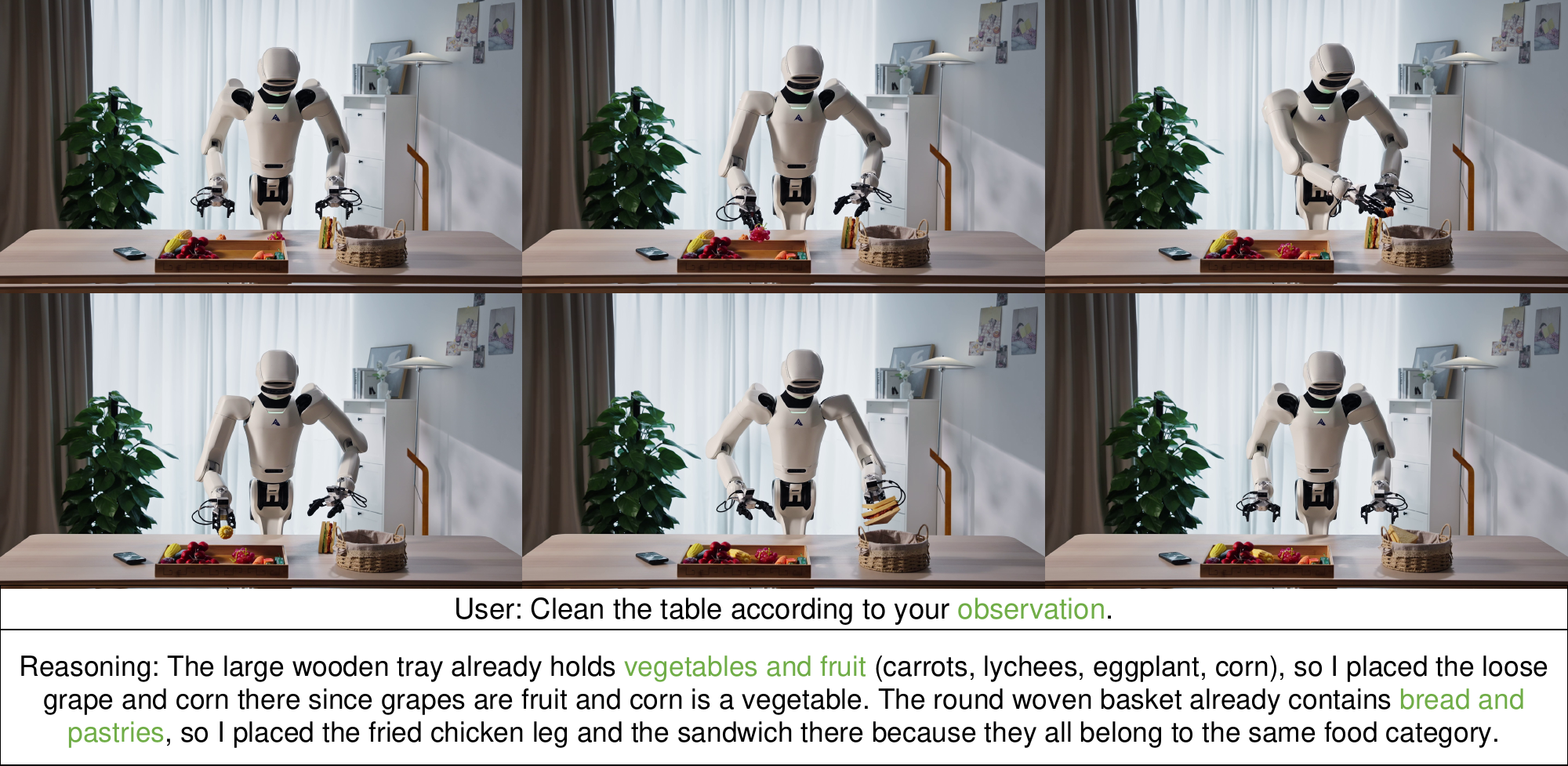}
        \label{fig:pattern}
    \end{subfigure}

    \vspace{-0.9em}

    \begin{subfigure}{\linewidth}
        \centering
        \caption{Preference-aware reasoning.}
        \includegraphics[width=\linewidth]{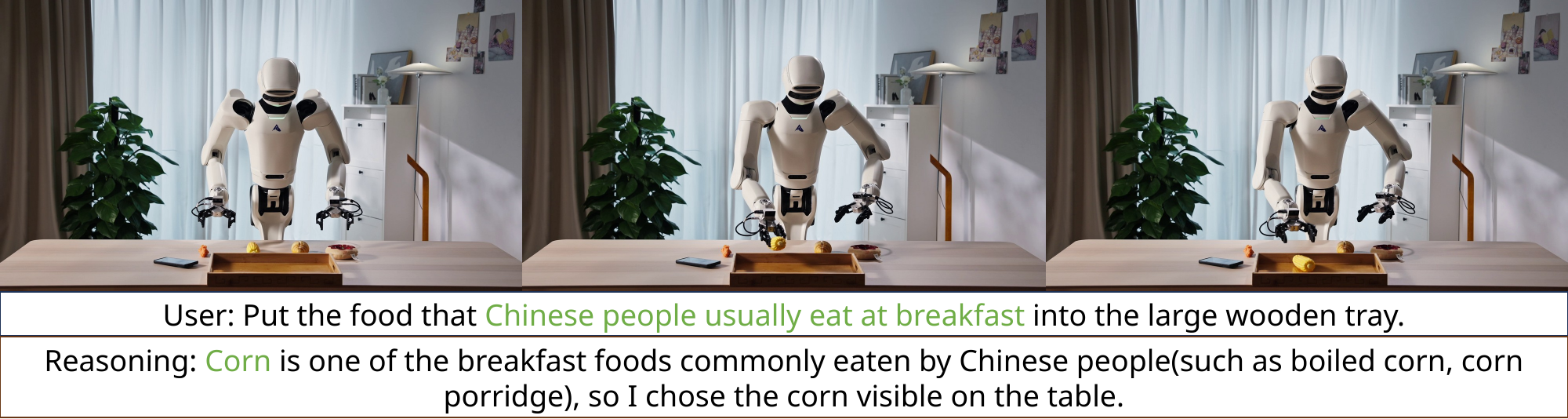}
        \label{fig:breakfast}
    \end{subfigure}

    \vspace{-0.9em}

    \begin{subfigure}{\linewidth}
        \centering
        \caption{High-calorie reasoning.}
        \includegraphics[width=\linewidth]{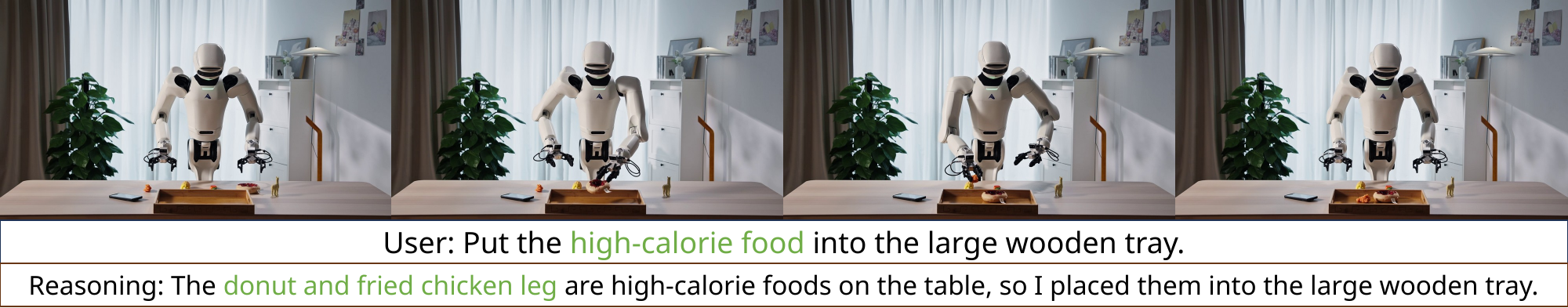}
        \label{fig:calorie}
    \end{subfigure}

    \vspace{-0.9em}

    \begin{subfigure}{\linewidth}
        \centering
        \caption{Food-category reasoning.}
        \includegraphics[width=\linewidth]{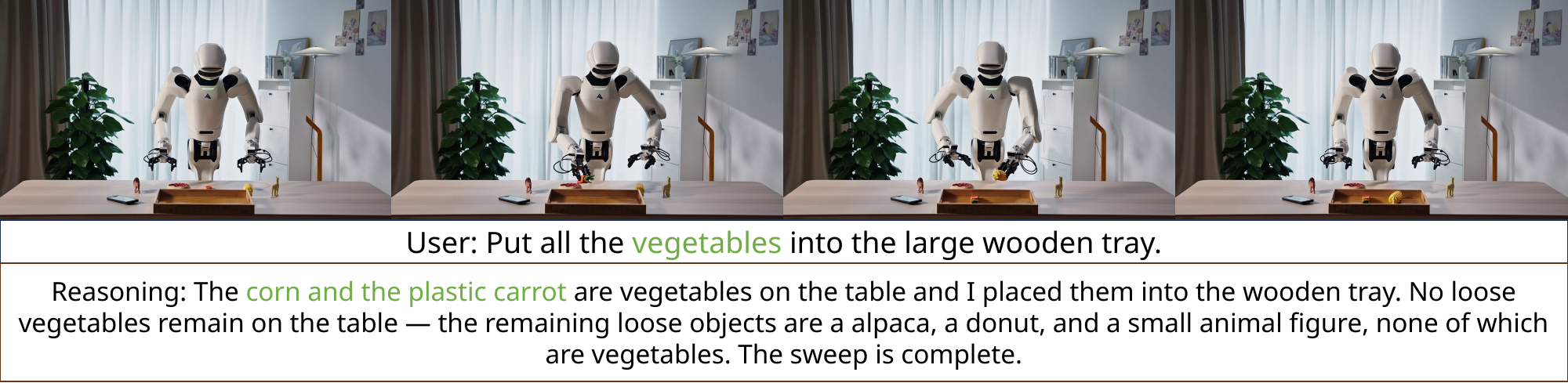}
        \label{fig:vegetable}
    \end{subfigure}

    \vspace{-1.0em}

    \caption{\textbf{Reasoning-grounded action.} \model transfers the agent's reasoning capability into policy execution. The examples demonstrate grounding in observations, user preferences, calorie knowledge, and food-category semantics.}
    \label{fig:reasoning}
\end{figure}

\subsection{PHILIA Playbook}

We validate \model through representative deployment scenarios on Astribot S1 robots. Each scenario exercises a distinct system capability: reasoning-grounded action, memory-grounded action, multi-robot coordination, or plug-and-play policy execution.

\paragraph{Agent-to-Policy Reasoning Grounding.} \model grounds high-level agent reasoning in low-level policy execution by constructing policy prompts from the user instruction, the robot's observation, and relevant external knowledge. As shown in Fig.~\ref{fig:pattern}, the agent combines scene observations with pretrained knowledge and web-search results before invoking a policy skill. This mechanism provides open-vocabulary grounding for policy execution and enables the robot to act on semantic constraints that are not hard-coded in the policy itself. Additional examples in Fig.~\ref{fig:breakfast}, Fig.~\ref{fig:calorie}, and Fig.~\ref{fig:vegetable} show reasoning grounded in user preferences, calorie knowledge, and food-category semantics.

\begin{figure}[H]
  \centering
  \includegraphics[width=\linewidth,keepaspectratio]{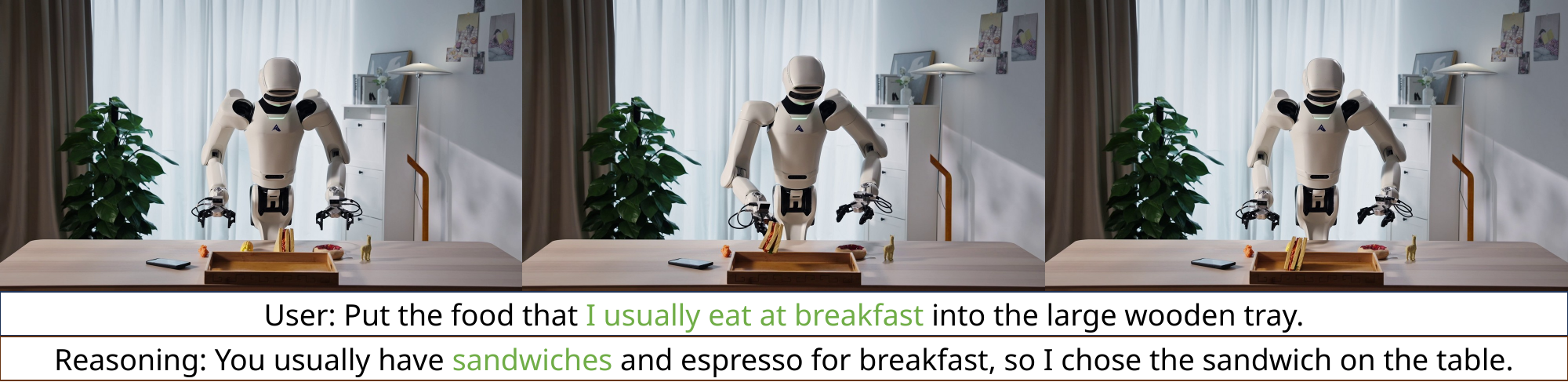}
   \caption{\textbf{Memory-grounded action.} \model retrieves long-term user preferences to guide related robot task execution.}
   \label{fig:memory}
\end{figure}

\paragraph{Agent-to-Policy Long-Term Memory Grounding.} \model also grounds physical action execution in long-term user memory. As shown in Fig.~\ref{fig:memory}, the user previously told \model that he usually has sandwiches and espresso for breakfast. \model stores this preference in memory and later retrieves it as context when handling related robot tasks.

\begin{figure}[H]
  \centering
  \includegraphics[width=\linewidth,keepaspectratio]{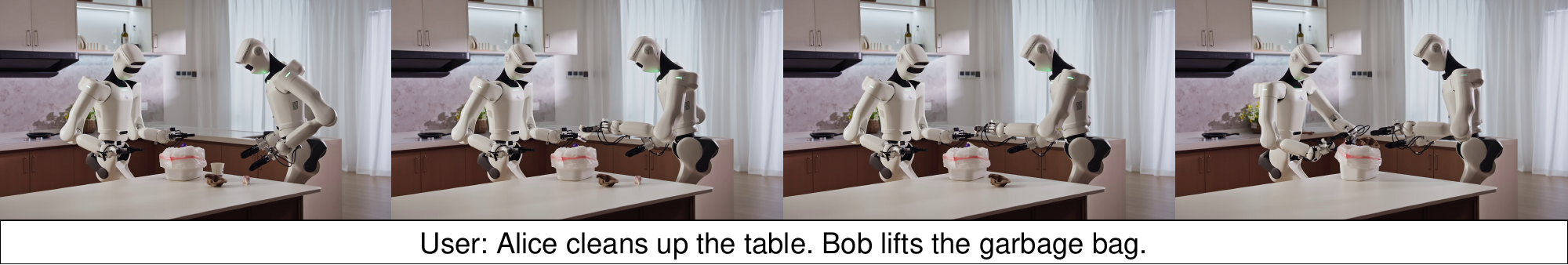}
   \caption{\textbf{One Agent, Multi-Robot Control.} \model controls multiple robots through independently registered robot gateways.}
   \label{fig:dual_robot}
\end{figure}

\paragraph{One Agent, Multi-Robot Control.} \model represents each robot through the same robot-gateway runtime abstraction. Adding a new robot therefore requires only instantiating a gateway and registering it with \model. As shown in Fig.~\ref{fig:dual_robot}, \model can control multiple robots by assigning tasks to their respective gateways in parallel, enabling robot Alice to clean up the table while robot Bob lifts the garbage bag.

\begin{figure}[H]
  \centering
  \includegraphics[width=\linewidth,keepaspectratio]{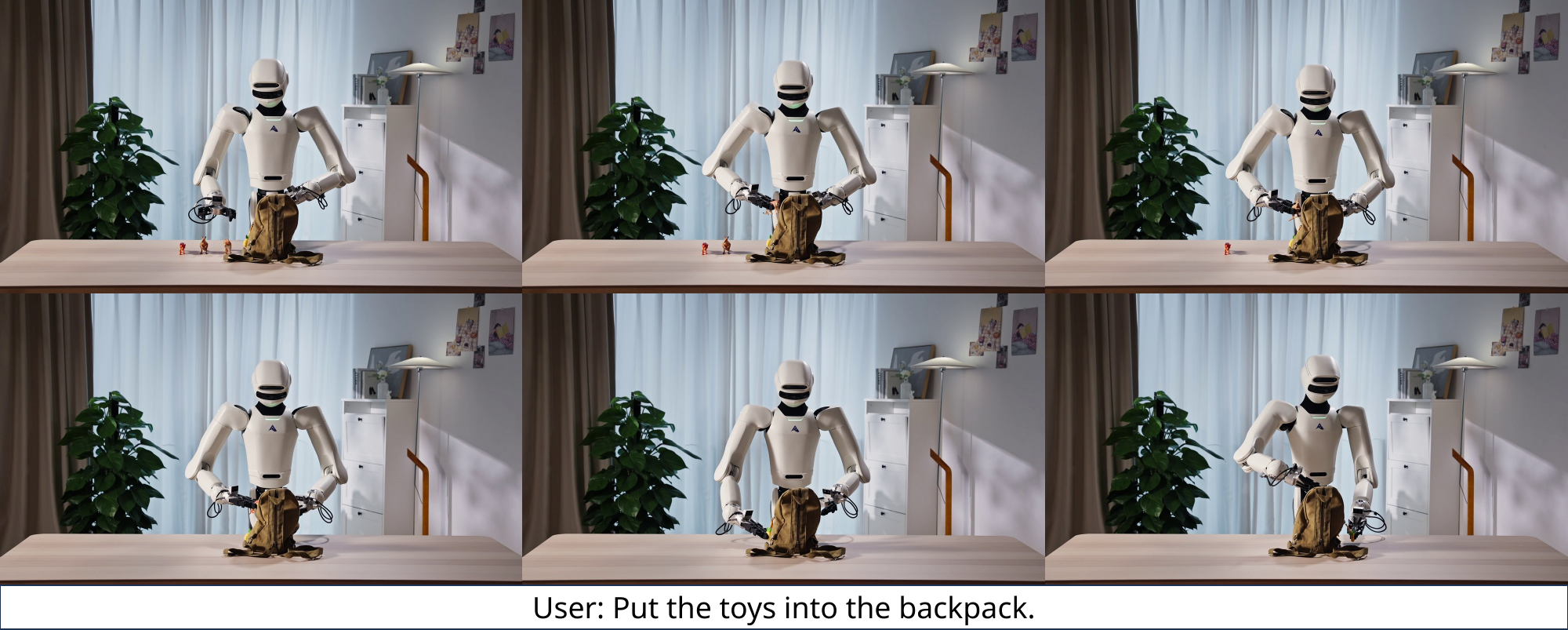}
   \caption{\textbf{Plug-and-play policy execution.} \model invokes a registered pack-the-backpack policy on the Lumo-2 robot.}
   \label{fig:backpack}
\end{figure}

\paragraph{Plug-and-Play Policy Execution.} \model treats each policy as a registered module, enabling plug-and-play policy execution across robots. Fig.~\ref{fig:backpack} shows \model invoking the Lumo-2 pack-the-backpack policy while preserving the robot's execution accuracy.

\section{Related Work}
\label{sec:related_work}

\paragraph{LLM-Based Robot Agents and Generalist Policies.}
Language-model-based robot systems have explored grounding instructions in affordances, tool calls, code, multi-modal representations, and spatial value maps, demonstrating that large models can provide high-level reasoning for embodied tasks~\citep{ahn2022saycan,liang2023codeaspolicies,driess2023palm,huang2023voxposer,autort2024}. More recent work extends these ideas to multi-robot task allocation, role assignment, and long-horizon collaboration~\citep{liu2023smartllm,sarkar2025llamar,zhang2025llmmrs}. In parallel, large-scale robot learning has produced generalist policies that connect language, vision, and action across robot embodiments and datasets~\citep{brohan2022rt,zitkovich2023rt,o2024open,team2024octo,kim2024openvla,black2024pi_0}, with recent work further exploring reasoning-aware action models, bimanual manipulation, and whole-body mobile manipulation~\citep{lumo1,liu2024rdt,fu2024mobile,zawalski2024robotic,zhao2025cot}. \model is complementary to these advances. Rather than proposing a new policy or asking the language model to directly control robots, it treats policies and robot skills as replaceable robot-local execution backends behind a shared gateway contract. The assistant remains responsible for intent interpretation, actor resolution, semantic memory, human-facing interaction, and recovery-level reasoning.

\paragraph{Long-Term Human--Robot Coexistence, Memory, and Personalization.}
Long-term service robots must operate in changing environments, preserve useful experience, and remain understandable to human users. The STRANDS project demonstrated extended autonomous deployments in everyday environments and highlighted the need to combine navigation, perception, interaction, planning, and learning for persistent service robots~\citep{hawes2017strands,kunze2018longterm}. Continual-learning studies for robotics and home service robots further show that personalization requires repeated human interaction, semantic knowledge updates, and careful handling of forgetting and trust~\citep{lesort2019continual,ayub2024interactive}. Personalized assistants such as TidyBot demonstrate how language models can infer household preferences from experience~\citep{wu2023tidybot}. At the same time, modern mobile robots rely on mature mapping, localization, and navigation systems, including LiDAR-inertial SLAM, graph optimization, semantic mapping, and planner stacks~\citep{xu2022fastlio2,shan2020liosam,hess2016cartographer,thrun2005probabilistic,kuemmerle2011g2o,kavraki1996prm,macenski2020nav2,kostavelis2015semantic}. \model treats persistent memory and robot geometry as shared support infrastructure for long-term coexistence: semantic memory informs planning across users and robots, while each robot retains its own maps, localization, and navigation stack.

\paragraph{Runtime Abstractions for Robot Execution and Recovery.}
Recent systems increasingly treat robots as components of an agentic runtime rather than isolated policies. RoboClaw-style systems expose robot services and skills through assistant-facing interfaces~\citep{spinmatrixroboclaw2026}, RoboClaw integrates data collection, policy learning, execution monitoring, and recovery~\citep{li2026roboclaw}, while RoboAgent studies long-horizon capability composition in simulated environments~\citep{xu2026roboagent}. ABot-Claw further extends OpenClaw toward persistent cooperative robotic agents, combining a unified embodiment interface, cross-embodiment multimodal memory, and critic-based closed-loop feedback for heterogeneous robot coordination~\citep{huo2026abotclaw}. Robotic systems also often structure task execution with modular control abstractions such as behavior trees, state machines, and task-level monitors; behavior trees are widely used because they are reactive, modular, and analyzable for robot missions~\citep{colledanchise2018behavior,iovino2020survey,ghzouli2022behavior}. \model similarly views robot execution as more than a single policy invocation, but introduces a deployment-oriented control plane over heterogeneous robot actors. Platform-specific execution, safety, monitoring, and recovery remain robot-local responsibilities exposed through gateway capabilities and status signals, while the assistant observes progress, requests cancellation, asks for human confirmation, and triggers bounded recovery without assuming authority over continuous robot control.

\paragraph{Middleware and Heterogeneous Robot Integration.}
Robotic middleware such as ROS and ROS~2 provides composition of drivers, sensors, planners, and policies within a single robot~\citep{quigley2009ros,macenski2022ros2}. \model operates at a different layer. It assumes each robot already possesses a functioning middleware-level software stack and exposes only a narrow semantic interface through a gateway. This allows robots built on ROS, ROS~2, or proprietary software to share a common assistant-facing capability contract. Rather than replacing planners, policies, navigation systems, or interaction interfaces, \model provides the deployment abstraction that connects them, enabling heterogeneous robot embodiments to be orchestrated as persistent physical agents behind a unified assistant identity. This separation makes actor-scoped routing, backend binding, capability discovery, stop/cancel ownership, and plug-in robot backends first-class system abstractions rather than implementation details.

\section{Discussion}
\label{sec:discussion}

\subsection{Core Design Tradeoffs}
The architecture is built around two deliberate tradeoffs that shape all higher-level design decisions:
\begin{enumerate}
    \item \textbf{Stable boundary vs. Granular state access.} \model prioritizes a narrow, stable gateway interface over full exposure of robot internals. This enables composable evolution across layers, but limits the agent's access to low-level execution state. The agent's ability to recover from failures (ambiguous scenes, blocked navigation, policy errors) depends entirely on the quality of semantic summaries published through the gateway contract.
    \item \textbf{Semantic flexibility vs. Physical safety.} The agent retains full flexibility for intent understanding, preference reasoning and task decomposition, but all state-changing physical actions pass through actor-scoped capabilities and safety gates. This preserves natural interaction richness without granting the language model direct authority over high-frequency motion control.
\end{enumerate}

Consistent with this boundary design, \model does not aim to replace underlying VLA policies, navigation stacks or robot middleware; instead, it provides a unifying runtime contract to compose them under a single assistant identity. This modular, boundary-driven design also determines the system's evolution pattern: user experience improves incrementally via upgrades to any individual layer (UI, policy, navigation, memory, dialogue), rather than relying on monolithic end-to-end model replacement. The gateway contract must therefore balance expressiveness and long-term stability.

\subsection{Heterogeneous Platform Support and Security}
The architecture natively supports integration of heterogeneous robot platforms beyond the currently validated Astribot S1. New robot models (e.g., Astribot T1) can be added by implementing a gateway adapter and publishing a capability manifest, with no changes to the agent control plane. This design property is not yet experimentally validated on non-S1 platforms, and remains a forward-looking feature of the architecture.

For physical security, authorization, confirmation, active-task arbitration and actor-scoped stop/cancel are built into the core system, not treated as optional UI features. The gateway boundary restricts the agent to structured capability requests, with platform credentials, map data and low-level safety checks retained exclusively in robot-local runtimes, narrowing the authority surface for physical execution.

\subsection{Limitations and Future Work}
This work focuses on architectural design and representative real-world use cases, not large-scale controlled benchmarking. The end-to-end task performance of manipulation, navigation and alignment depends on the quality of the underlying robot-local backends, which are treated as pluggable components.

Key open directions for future work include: quantifying long-term reliability and failure recovery rates in sustained daily operation; evaluating multi-user authorization and permission management; and validating cross-platform generalization across heterogeneous robot embodiments beyond S1. We also note that current gateway semantics remain relatively coarse-grained, and finer-grained state feedback could further improve agent-level failure handling.

\section{Conclusion}
\label{sec:conclusion}

We presented \model, a system architecture for long-term physical coexistence with intelligent robots. The central idea is to keep a persistent assistant identity at the semantic control plane while exposing robot platforms through simple robot gateway abstractions. Rather than replacing underlying VLA policies, navigation stacks or robot middleware, \model provides a unifying runtime contract to compose them under a single assistant identity. This design preserves rich interaction, tools, UI surfaces, memory, and planning from OpenClaw, while delegating fine-tuned policies, navigation, sensing, and safety-critical execution to robot-local runtimes.

New robot platforms can be integrated by implementing a gateway adapter and publishing a capability manifest, with no modifications to the agent control plane. Through representative use cases, \model illustrates how agent-level reasoning and long-horizon context can be grounded into physical robot action without tightly coupling the assistant to a single body, robot model, or policy backend.

Instead of replacing human agency, judgment and responsibility, systems like \model extend human capability by distributing cognition across people, machines and environments. This work is an invitation to explore how intelligent robots can become reliable, personalized partners in long-term daily coexistence.

\newpage

\section{Contributions}
Author contributions in the following areas are listed in alphabetical order.
\label{sec:contribution}
\begin{itemize}
    \item \textbf{System Design:} Weiqi Jin
    \item  \textbf{Contributors:} Baifu Huang, Binyan Sun, Haotian Yang, Kuncheng Luo, Peijun Tang, Shangjin Xie
    \item  \textbf{Project Lead:} Jianan Wang

\end{itemize}

\clearpage

\bibliography{main}

\end{document}